\documentclass[conference]{IEEEtran}
\IEEEoverridecommandlockouts

\usepackage{cite}
\usepackage{amsmath,amssymb,amsfonts}
\usepackage{algorithmic}
\usepackage{graphicx}
\usepackage{textcomp}
\usepackage[table]{xcolor}
\usepackage{multirow}
\usepackage{booktabs}
\usepackage{subcaption}
\usepackage{url}
\usepackage[hidelinks]{hyperref}

\newcommand{\revised}[1]{#1}

\def\BibTeX{{\rm B\kern-.05em{\sc i\kern-.025em b}\kern-.08em
    T\kern-.1667em\lower.7ex\hbox{E}\kern-.125emX}}

\newlength{\authcolwidth}
\newcommand{\authblock}[2]{%
  \IEEEauthorblockN{\makebox[\authcolwidth]{#1}}%
  \IEEEauthorblockA{\parbox[t]{\authcolwidth}{\centering #2}}}

\selectfont

\begin{document}

\title{MoCo-AIS: A Contrastive Learning Framework for\\Similarity Computation of Vessel Trajectories}

\author{%
\authblock{Ruixin Song}{%
  \textit{Faculty of Computer Science} \\
  \textit{Dalhousie University}\\
  Halifax, Canada \\
  rsong@dal.ca}
\and[]
\authblock{Md Mahbub Alam}{%
  \textit{Faculty of Computer Science} \\
  \textit{Dalhousie University}\\
  Halifax, Canada \\
  mahbub.alam@dal.ca}
\and[]
\authblock{Zahra Sadeghi}{%
  \textit{Faculty of Computer Science} \\
  \textit{Dalhousie University}\\
  Halifax, Canada \\
  zahras@dal.ca}
\and[\linebreak]
\authblock{Amilcar Soares}{%
  \textit{Computer Science and Media Technology} \\
  \textit{Linnaeus University}\\
  Växjö, Sweden \\
  amilcar.soares@lnu.se}
\and[]
\authblock{Jose F. Rodrigues-Jr}{%
  \textit{Computer Science Department} \\
  \textit{University of São Paulo}\\
  São Carlos, Brazil \\
  junio@icmc.usp.br}
\and[]
\authblock{Gabriel Spadon$^{\star}$}{%
  \textit{Faculty of Computer Science} \\
  \textit{Dalhousie University}\\
  Halifax, Canada \\
  spadon@dal.ca}
\thanks{$\star$~Corresponding author.}
}

\maketitle

\begin{abstract}
Trajectory similarity is a fundamental task in analyzing mobility patterns, essential for applications such as route pattern extraction, mobility prediction, and anomaly detection. Traditional distance-based measures for computing similarity incur high computational cost, driving the adoption of lightweight learning-based approaches. Supervised methods rely on extensive labels derived from traditional distance measures and often reproduce these metrics, which limits generalization. Self-supervised contrastive learning addresses this issue but lacks a unified framework, making it difficult to compare deep learning (DL) models for consistent trajectory representation. This paper presents MoCo-AIS, a unified framework for learning vessel trajectory embeddings based on the Momentum Contrast (MoCo) paradigm, which formulates similarity learning through positive and negative trajectory pairs. Within it, we evaluate leading DL models on large-scale, real-world vessel-tracking AIS datasets covering diverse navigation behaviors and operating conditions. Results show that MoCo-AIS improves trajectory retrieval and achieves the best retrieval performance among the learned models under a fixed input representation, while reducing large-scale similarity computation and providing a unified benchmarking platform for trajectory representation models.
\end{abstract}

\begin{IEEEkeywords}
AIS data, trajectory similarity, representation learning, contrastive learning, spatio-temporal systems
\end{IEEEkeywords}

\section{Introduction}
A trajectory represents the motion of an object in space over time. It is typically defined as an ordered sequence of observations, each associated with spatial coordinates, temporal information, and dynamic features. Trajectories are collected from diverse sources, including GPS-enabled devices, sensor networks, and the Automatic Identification System (AIS), capturing the movements of humans, vehicles, animals, and natural phenomena such as hurricanes and tornadoes~\cite{zheng2015trajectory}. The widespread availability of trajectory data and its integration into location-based services have made trajectory analysis increasingly important in academia and industry, with dual-use relevance in both civilian and security domains~\cite{alam2022survey, fournier2018past}.

Since 2004, AIS has been mandatory for commercial vessels, enhancing maritime safety and security by continuously broadcasting positional and voyage-related information~\cite{Tu2018ExploitingAISData}. The large scale of historical AIS data has enabled extensive research on vessel movement patterns, including route extraction, weather-aware routing~\cite{Troupiotis-Kapeliaris2025DynamicWeatherResilientVessel}, anomalous behavior detection~\cite{Sadeghi2023AnomalyDetectionMaritime}, and mobility prediction~\cite{troupiotis2025vessel}. Central to many of these analyses is trajectory similarity, which quantifies how closely two paths align across spatial, temporal, and semantic dimensions~\cite{chang2023trajectory}. This notion supports many downstream tasks: clustering trajectories~\cite{Alam2022Clustering}, searching for similar paths~\cite{chen2005robust}, trajectory classification~\cite{westerdijk2019classifying}, identifying co-movement patterns~\cite{Ferrero2018MOVELETSExploringRelevant, Ferrero2020MasterMoveletsDiscoveringHeterogeneous, TortelliPortela2022HiPerMoveletsHighperformanceMovelet}, and extracting mobility networks~\cite{huang2024generation, spadon2019reconstructing}.

Measuring similarity for vessel trajectories is harder than for land-based GPS data: vessels operate in loosely bounded environments where weather (\textit{e.g.}, wind and currents) and regional regulations influence movements, causing positional uncertainty and frequent maneuver adjustments~\cite{Alam2024EnhancingShorttermVessel}. AIS messages are also often irregular over long voyages~\cite{Liang2024SurveyDistanceBasedVessel} due to burst transmissions, dropouts, and uneven coverage~\cite{Tu2018ExploitingAISData}.

Distance-based methods (\textit{e.g.}, Hausdorff~\cite{hausdorff1914} and Dynamic Time Warping (DTW)~\cite{Keogh2005ExactIndexingDynamic}) measure trajectory similarity by aligning raw coordinate sequences and computing point-wise distances, capturing mainly geometric similarity~\cite{su2020survey, tao2021comparative}.
They are widely used in mobility studies for tasks such as clustering vessel routes~\cite{Alam2022Clustering} and predicting vessel movements~\cite{suo2020ship, Alam2024EnhancingShorttermVessel}. However, as trajectory datasets grow, distance-based methods face increasing computational bottlenecks.

Although computationally expensive, distance-based approaches remain widely used in maritime studies due to their well-defined mathematical properties~\cite{Alam2022Clustering, chen2005robust, westerdijk2019classifying, huang2024generation}. Yet, the rapid growth of AIS data underscores the pressing need for methods that achieve comparable accuracy at lower computational cost. Recent learning-based approaches have begun addressing this challenge using deep neural networks, but they rely on extensive labeling with traditional distance metrics, and models trained on a specific metric reproduce it by construction~\cite{Yao2019ComputingTrajectorySimilarity, Yang2021T3SEffectiveRepresentation}.
This limits progress by constraining learning-based studies to approximating existing methods rather than advancing toward generalizable trajectory embeddings.

In contrast, self-supervised methods leverage contrastive learning to learn discriminative embeddings that preserve relational structures among trajectories without manual labeling~\cite{Li2018DeepRepresentationLearning, Liu2022CSTRMContrastiveSelfSupervised, Chang2023ContrastiveTrajectorySimilarity}. Unlike supervised approaches that replicate predefined measures or reconstruction-based models focusing on point-level AIS restoration~\cite{Zhang2018NovelShipTrajectory, Chen2025OptimizingVesselTrajectories}, contrastive learning infers similarities through data augmentations~\cite{Liu2022CSTRMContrastiveSelfSupervised, Chang2023ContrastiveTrajectorySimilarity} or distance-guided sample swapping strategies~\cite{Li2018DeepRepresentationLearning}. This capability is especially valuable for AIS data, where global similarity labels are inherently ambiguous~\cite{Ferreira2022SemiSupervisedMethodologyFishing, Liang2024SurveyDistanceBasedVessel}.
Yet most studies prioritize new architectures and marginal accuracy gains over consistent evaluation frameworks, leading to heterogeneous experimental setups (\textit{e.g.}, baseline architectures, datasets, preprocessing pipelines, and metrics) that obscure how methodological choices influence similarity learning.

In this study, we addressed trajectory similarity by learning embedding functions that map vessel trajectories into a representation space where distances reflect meaningful spatio-temporal similarity. We introduced MoCo-AIS, a unified framework that adapts Momentum Contrast (MoCo)~\cite{He2020MomentumContrastUnsupervised} to AIS data by coupling trajectory-specific augmentations with interchangeable deep encoder architectures designed to capture spatial dynamics, temporal dependencies, and motion patterns. Within it, contrastive learning pulls embeddings of semantically similar segments together and pushes dissimilar ones apart, enabling similarity measurement directly in the learned space. We adopted MoCo because its momentum encoder and dynamic memory dictionary provide a large, consistent set of negative samples without large batch sizes, improving computational efficiency and training stability. Our main contributions are as follows:

\begin{itemize}
    \item We proposed a unified framework for learning vessel-trajectory embeddings for similarity computation and benchmark various deep learning architectures.

    \item We standardized the evaluation protocol on similarity search, comparing the unified framework against (i) traditional similarity measures (Hausdorff and DTW) and (ii) self-supervised baselines (t2vec and TrajCL).

    \item We evaluated the learned representation for robustness (across held-out trajectory perturbations) and transferability (across different geographical regions).

    \item We benchmarked computational efficiency by computing pairwise similarity matrices from learned embeddings and classical distance-based measures, demonstrating more effective and scalable trajectory comparison.
\end{itemize}

The rest of the paper is organized as follows. Section~\ref{sec:related-work} reviews recent advances in trajectory similarity measurement. Section~\ref{sec:def-and-notation} presents the preliminaries and notation. Section~\ref{sec:methodology} introduces the proposed MoCo-AIS framework for learning trajectory embeddings. Section~\ref{sec:exp-results} reports the experimental evaluation and analysis based on real-world AIS datasets. Section~\ref{sec:conclusion} concludes with future research directions.

\section{Related Work}
\label{sec:related-work}

\subsection{Distance-based Methods}

Distance-based methods compare trajectories by geometry and alignment. The Hausdorff distance~\cite{hausdorff1914} measures the maximum spatial deviation between trajectories but is outlier-sensitive. DTW~\cite{Keogh2005ExactIndexingDynamic} warps sequences to minimize cumulative distance, accommodating unequal lengths at high computational cost and with similar noise sensitivity.

Beyond spatial distance, Multidimensional Similarity Measures enrich comparison with semantic features~\cite{Furtado2016MultidimensionalSimilarityMeasuring}; a hierarchical partitioning tree over trajectory frequency vectors scores similarity by node depth~\cite{Varlamis2021NovelSimilarityMeasure}, and indexing over high-dimensional features enables efficient retrieval~\cite{Souza2022MATIndexIndexFast}.

These methods produce interpretable scores but are computationally intensive and sensitive to sampling density. We address both limitations by learning latent trajectory embeddings, enabling efficient similarity retrieval.

\subsection{Learning-based Methods}
\label{subsec:learning-methods}

\subsubsection{Supervised Methods}

Supervised similarity learning labels pairs with distances from a chosen measure (\textit{e.g.}, DTW, longest common subsequence, or Hausdorff) and trains models to predict them, using recurrent neural network (RNN)-based (NeuTraj~\cite{Yao2019ComputingTrajectorySimilarity}), convolutional neural network (CNN)-based (TrajSR~\cite{Cao2021AccurateComputationTrajectory}), and graph attention-based (TrajGAT~\cite{Yao2022TrajGATGraphbasedLongterm}) architectures. Most open-source trajectory data is unlabeled, however, and annotating it is computationally expensive; labels from one measure bias the embeddings toward it.

\subsubsection{Self-supervised Methods}

Self-supervised methods learn latent embeddings that preserve relational distances among trajectories, typically through contrastive or triplet objectives. t2vec~\cite{Li2018DeepRepresentationLearning} tokenizes trajectory sequences and applies stacked GRUs with a triplet loss~\cite{WeinbergerDistanceMetricLearning}, while TrajCL~\cite{Chang2023ContrastiveTrajectorySimilarity} builds on MoCo~\cite{He2020MomentumContrastUnsupervised}, treating augmented samples as positives and drawing negatives from a dynamic memory bank under the InfoNCE loss~\cite{Oord2019RepresentationLearningContrastive}. CLEAR~\cite{Li2024CLEARRankedMultiPositive} refines the positive side of this construction, generating several variations of an anchor trajectory and ranking them by their similarity to it so that harder positives are weighted differently, which improves robustness to noise and to low sampling rates.

Encoder design has drawn increasing attention: recurrent architectures remain widely used for sequential modeling~\cite{Li2018DeepRepresentationLearning, Zhang2020TrajectorySimilarityLearning}, convolutional encoders capture multi-scale spatial patterns~\cite{Li2025HiTJEPAHierarchicalSelfsupervised}, and Transformers model long-range dependencies~\cite{Liu2022CSTRMContrastiveSelfSupervised, Chang2023ContrastiveTrajectorySimilarity}. Further directions add relational and geometric inductive biases through graph neural networks~\cite{Chen2024KGTSContrastiveTrajectory}, hypergraphs~\cite{Cao2024HypergraphHashLearning}, and non-Euclidean embedding spaces~\cite{Si2025RobustTrajectoryEmbedding}.

Supervised methods thus depend on expensive, measure-derived labels and may inherit their biases, whereas self-supervised methods learn directly from data but require careful augmentation and rigorous validation. Differences in architecture and objective across studies make fair comparison difficult, motivating the need for a unified evaluation framework.

\subsection{Methods for Vessel Trajectory Similarity}

Most learning-based research targets urban traffic, where road networks constrain mobility; maritime traffic has received less attention, and distance measures remain dominant in vessel mobility analysis. Nie \textit{et al.}~\cite{Nie2021TrajectorySimilarityAnalysis} construct buffers around AIS trajectories and measure proximity to nearby traces; Liang \textit{et al.}~\cite{Liang2021UnsupervisedLearningMethod} apply convolutional autoencoders to grid-based images; AISim~\cite{Luo2023VesselTrajectorySimilarity} embeds AIS trajectories with heterogeneous graph neural networks supervised by distance measures; and ACTIVE~\cite{Liu2025ACTIVEContinuousSimilarity} combines segment-based indexing with continuous similarity search for real-time retrieval.

Key limitations therefore remain for AIS trajectory similarity: distance-based methods are interpretable but scale poorly and are sensitive to irregular sampling; supervised approaches inherit bias and cost from their reference measures; and self-supervised models remove labels but their architectural and protocol diversity hinders fair comparison. Many maritime studies also import assumptions from road traffic, ignoring the sparsity, long-range dependencies, and environmental factors of AIS data. We address these issues with MoCo-AIS, a unified self-supervised framework tailored to maritime trajectories, together with a standardized evaluation protocol.

\section{Preliminaries and Notation}
\label{sec:def-and-notation}

\noindent\textit{\textbf{Trajectory}}.
An AIS trajectory $\mathcal{T} = \{p_1, p_2, \dots, p_L\}$ is a sequence of $L$ ordered points produced by a moving vessel over time, where each point $p_t$ in the trajectory represents the object's coordinates and kinematic state at time $t$, \textit{i.e.}, $p_i = (x_i, y_i, v_i, \psi_i, t_i)$, where:
\begin{itemize}
    \item $x_i$: Longitude (coordinate, in degrees, $[-180^\circ, 180^\circ]$);
    \item $y_i$: Latitude (coordinate, in degrees, $[-90^\circ, 90^\circ]$);
    \item $v_i$: Speed Over Ground (SOG, in knots);
    \item $\psi_i$: Course Over Ground (COG, in degrees, $[0, 360]$); and,
    \item $t_i$: the transmission timestamp (standard Unix).
\end{itemize}
In this study, we restricted similarity computation to the spatial sequence $(x_i, y_i)_{i=1}^L$ while preserving temporal order, focusing on geometric path similarity and ensuring comparable evaluation across distance-based and learning-based methods.

\noindent\textit{\textbf{Trajectory Dataset}}.
Let $D=\{\mathcal{T}_1,\dots,\mathcal{T}_n\}$ denote $n$ trajectories, where $\mathcal{T}_i$ is a variable-length sequence of points.

\noindent\textit{\textbf{Trajectory Similarity}}.
For trajectories $\mathcal{T}_i, \mathcal{T}_j \in D$, a \textit{non-learned similarity} function $f_n$ takes $\mathcal{T}_i$ and $\mathcal{T}_j$ as input and returns a similarity score that quantifies their spatial correspondence. Let $\mathrm{dist}(\mathcal{T}_i,\mathcal{T}_j)$ denote a trajectory distance computed using a pointwise distance (\textit{i.e.}, Euclidean). Then, similarity is defined as the complement of distance:
\begin{equation}
   f_n(\mathcal{T}_i, \mathcal{T}_j) = 1 - \mathrm{dist}(\mathcal{T}_i, \mathcal{T}_j)
\end{equation}

A \textit{learned similarity} function $f_l$ instead compares feature representations of trajectories in a latent embedding space. This process involves an encoder,
$
f_{enc}: \mathcal{T} \rightarrow \mathbb{R}^d,
$
typically a sequence model such as a long short-term memory (LSTM) network, whose parameters $\theta$ are optimized by supervised or self-supervised learning. Similarity between two trajectories is then computed by a predefined metric (\textit{e.g.}, cosine similarity) over learned embeddings:
\begin{equation}
f_l(\mathcal{T}_i, \mathcal{T}_j) = \text{sim} \big(f_{enc}(\mathcal{T}_i;\theta), f_{enc}(\mathcal{T}_j;\theta)\big)
\end{equation}

\noindent\textit{\textbf{Contrastive Learning}}.
A representation learning paradigm that trains an encoder $f_{enc}$ to map trajectories into a latent embedding space, where similar trajectories are placed in proximity and dissimilar ones are pushed farther apart. Given a trajectory $\mathcal{T}_i$ in dataset $D$, a positive sample $\mathcal{T}_i^+$ is generated through augmentation, and the other trajectory $\mathcal{T}_k\ (k\ne i)$ in $D$ is considered a negative sample. During training, the encoder $f_{enc}$ is optimized using a contrastive objective. Formally, each trajectory $\mathcal{T}_i$, its positive counterpart $\mathcal{T}_i^+$, and the negative sample set ${\mathcal{T}_k}^-$ are mapped to embeddings as:
\begin{equation}
\begin{aligned}
    \mathbf{z}_i &= f_{enc}(\mathcal{T}_i), \\
    \mathbf{z}_i^{+} &= f_{enc}(\mathcal{T}_i^{+}), \\
    \mathcal{Z}_i^{-} &= \{\mathbf{z}_i^{-}\} = \{\, f_{enc}(\mathcal{T}_k) \mid \mathcal{T}_k \in D_i^- \subseteq D,\, k \ne i \,\}.
\end{aligned}
\end{equation}
The optimization objective, a contrastive loss function (\textit{e.g.}, InfoNCE), minimizes the distance between $\mathbf{z}_i$ and $\mathbf{z}_i^+$, while maximizing the one between $\mathbf{z}_i$ and $\mathcal{Z}_i^-$ embeddings.

\section{Methodology}
\label{sec:methodology}

\begin{figure}[!htbp]
    \centering
    \includegraphics[width=0.76\columnwidth]{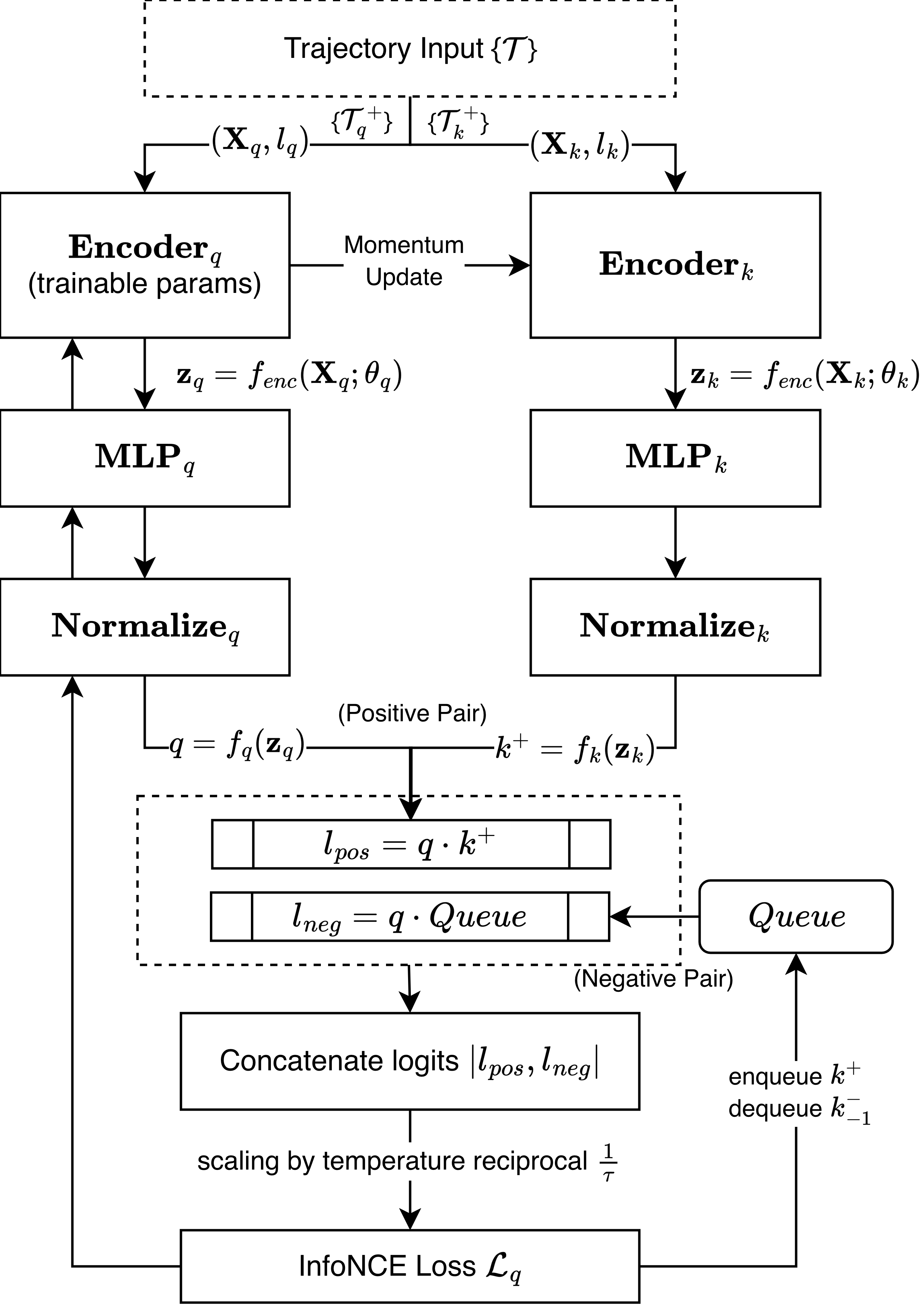}
    \caption{MoCo-AIS architecture for vessel trajectory similarity.}
    \label{fig:framework}
\end{figure}

MoCo-AIS is a framework for learning trajectory-similarity representations based on momentum-contrastive learning~\cite{He2020MomentumContrastUnsupervised}. While TrajCL~\cite{Chang2023ContrastiveTrajectorySimilarity} applied MoCo-inspired structures to terrestrial trajectories using Transformer backbones, MoCo-AIS generalized this principle through a modular design that housed multiple sequential modeling architectures. It introduced a flexible encoder plugin system supporting RNN-based models, \textit{i.e.}, LSTMs and gated recurrent units (GRUs), and Temporal Convolutional Networks (TCNs). As shown in Fig.~\ref{fig:framework}, the framework comprised query and key encoders with momentum updates, projection heads, a First-In--First-Out (FIFO) queue of negative samples, and an InfoNCE loss, with details on the components and design choices used to process raw coordinate sequences.

\subsection{Trajectory Input}

Vessel movement patterns are inherently heterogeneous, with mobility modes often shaped by abrupt maneuvers driven by environmental conditions such as strong winds, currents, and adverse weather. Although speed over ground and course over ground were initially considered alongside latitude and longitude, these kinematic attributes exhibit high variability due to local noise and irregular sampling. As a result, they were excluded from the final representation. The learning process, therefore, relied solely on geographic coordinates $(x_i, y_i)$ and their temporal order, preserving the spatial structure of vessel trajectories while avoiding noise amplification. In this paper, each trajectory is an ordered sequence of coordinate pairs (\ref{eq:traj-pair}), where $L$ is its length.
\begin{equation}
    \label{eq:traj-pair}
    \mathcal{T} = \left[ (x_1, y_1), (x_2, y_2), (x_3, y_3), \dots, (x_L, y_L) \right] \in \mathbb{R}^{L \times 2}
\end{equation}

\noindent\textit{\textbf{Similar Samples}}.
For each trajectory $\mathcal{T}$, two augmented views were generated (the query $\mathcal{T}_q^+$ and the key $\mathcal{T}_k^+$), which together form a positive pair for contrastive learning. We employed three augmentation strategies~\cite{Liu2022CSTRMContrastiveSelfSupervised, Chang2023ContrastiveTrajectorySimilarity}. The first two were applied during contrastive training, while the third was held for evaluation:
\begin{enumerate}
    \renewcommand{\labelenumi}{(\theenumi)}
    \item sub-trajectory, which removes a subset of points;
    \item shape distortion (shifting), which perturbs point positions; and
    \item simplification with the Ramer-Douglas-Peucker algorithm~\cite{DOUGLAS1973AlgorithmsReductionNumber}, which reduces trajectory complexity while preserving its overall structure.
\end{enumerate}




Trajectory sequences within a batch can vary in length. For efficient batching, trajectories $\mathcal{T}_q^+$ and $\mathcal{T}_k^+$ were padded to a uniform maximum length $T$ within each batch, using a binary matrix mask to distinguish valid and padded positions. The batched input is a tuple $((\mathbf{X}_q, l_q), (\mathbf{X}_k, l_k))$, where $\mathbf{X} \in \mathbb{R}^{B \times T \times 2}$ denotes a padded batch of trajectories, $B$ is the batch size, $T$ is the padded length and $l$ is a vector of true sequence lengths. This tuple provides the standardized query and key inputs to the model (see Fig.~\ref{fig:framework}), ensuring compatibility with different encoders as plugins in the framework.

\subsection{Encoder}
\label{subsec:encoders}

Our framework supported a range of encoders that could be integrated, with both key and query encoders sharing the same architecture. To accommodate variable-length sequences, all modules employed a consistent masking and pooling strategy.
\begin{figure}[!htbp]
    \centering
    \includegraphics[width=0.55\columnwidth]{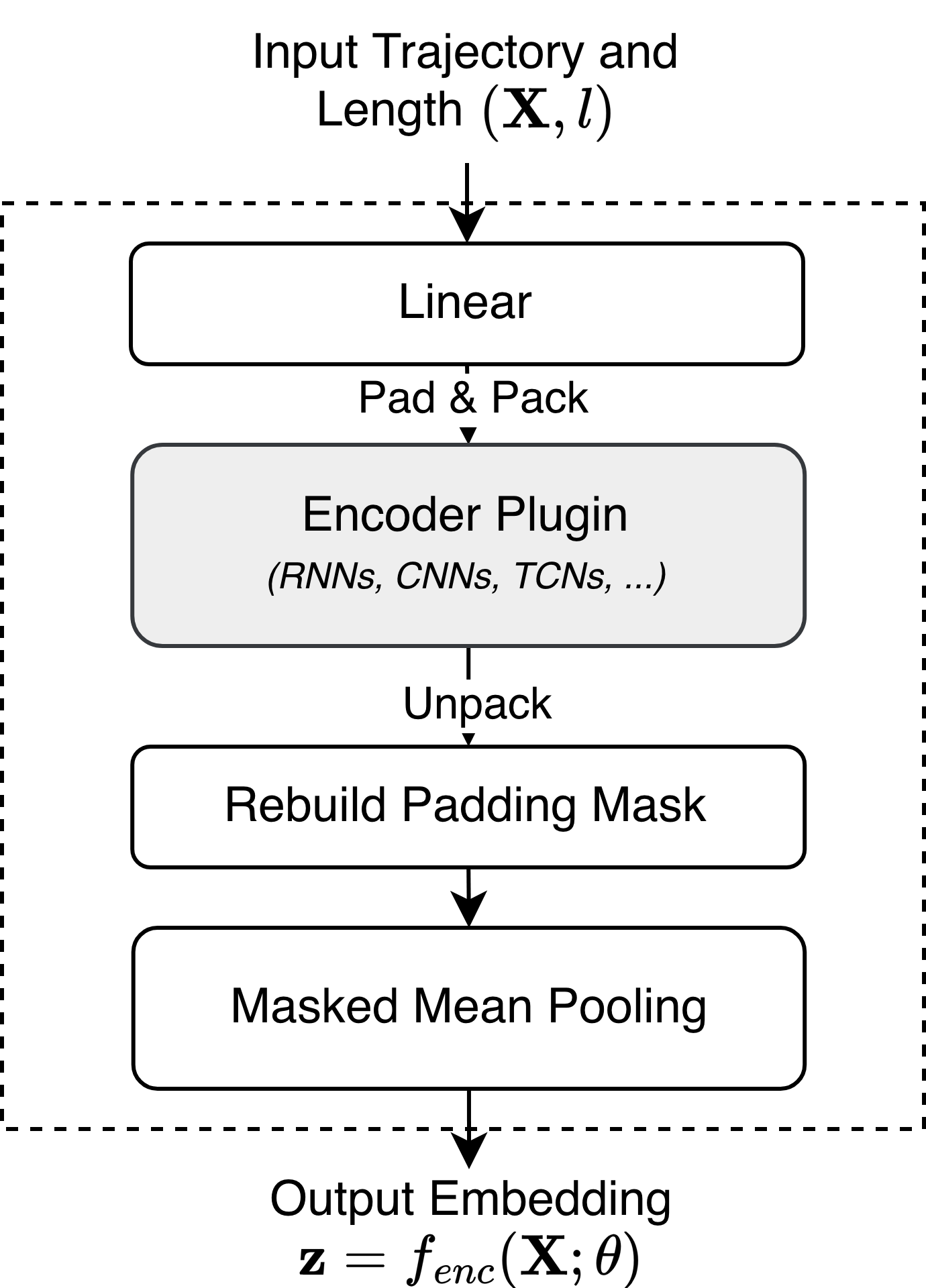}
    \caption{Encoder plugin supported by the framework.}
    \label{fig:encoder}
\end{figure}

\subsubsection{RNN- and Convolution-based Encoders}
Fig.~\ref{fig:encoder} illustrates encoders based on recurrent and convolutional models, including LSTMs, GRUs, and TCNs. Input points were linearly projected into a $d$-dimensional space and packed according to their actual sequence lengths. The packed sequences were then fed into a stack of GRU/LSTM/TCN units. The output states were processed using masked mean pooling to generate the final embeddings. In our experiments, we evaluated bidirectional GRUs/LSTMs and TCNs.

\subsubsection{Padding and Masked Mean Pooling}
Each input trajectory in a batch was padded to length $T$, with a padding mask $M$ indicating padded and valid positions to prevent tokens from contributing false information. Given hidden states $\mathbf{H}$ output from the encoder, we applied masked mean pooling to obtain a fixed-size embedding for each trajectory:
\begin{equation}
    \mathbf{z}_i = \frac{1}{L_i}\sum_{t=1}^{T}(1-M_{i,t})\cdot \mathbf{H}_{i,t}
\label{eq:mask-pool}
\end{equation}
where $L_i$ is the true sequence length, so only valid points contribute to the final embedding and padding is ignored.

\subsection{Loss Function, Memory Queue, and Momentum Update}
\label{subsec:contrastive}

\subsubsection{Memory Queue Update}
For negative examples, the framework maintained a fixed-size memory queue that stored encoded trajectory embeddings. After each training step, the current key embedding $k_t$ was enqueued, and the oldest entry $k_{t-K}$ was dequeued to preserve the queue length:
\begin{equation}
    Q_t = \left[ k_t, k_{t-1},...,k_{t-K+1}\right]
\end{equation}
The query embedding $q$ was contrasted with all entries in $Q_t$, while maximizing similarity with the positive embedding $k^+$.

\subsubsection{InfoNCE Loss}
The contrastive learning objective is the InfoNCE loss, which is closely related to the cross-entropy loss used in classification tasks. Given the query embedding $q$, a positive key embedding $k^+$, and the negative set $Q_t=\{k^-_i \}^K_{i=1}$, the loss is defined as:
\begin{equation}
    \mathcal{L}_q = -\log \frac{\exp(\mathrm{sim}(q,k^+)/\tau)}
    {\sum_{k \in \{k^+\} \cup Q_t} \exp(\mathrm{sim}(q,k)/\tau)}
    \label{eq:nceloss}
\end{equation}
where $\mathrm{sim}(\cdot,\cdot)$ denotes cosine similarity and $\tau$ controls the sharpness of the softmax distribution: smaller $\tau$ sharpens the distribution by focusing on the most similar samples, while larger $\tau$ produces a smoother distribution.

\subsubsection{Momentum Update}
Of the two encoders in Fig.~\ref{fig:framework}, the query parameters $\theta_q$ were updated by backpropagation and the key parameters $\theta_k$ by momentum from $\theta_q$:
\begin{equation}
\label{eq:momentum-update}
    \theta_k \xleftarrow{} m \times \theta_k + (1-m) \times \theta_q
\end{equation}
where $m\in [0,1)$ is the momentum coefficient. The key encoder was kept in inference mode, \textit{i.e.}, not updated by gradients, ensuring smooth parameter updates and consistent negative representations in the memory queue.

\section{Experiments and Analysis}
\label{sec:exp-results}

\subsection{Experiment Setup}

\begin{figure*}[!htbp]
\centering
\begin{subfigure}[b]{0.332\linewidth}
    \centering
    \includegraphics[width=\linewidth,trim=112 8 112 8,clip]{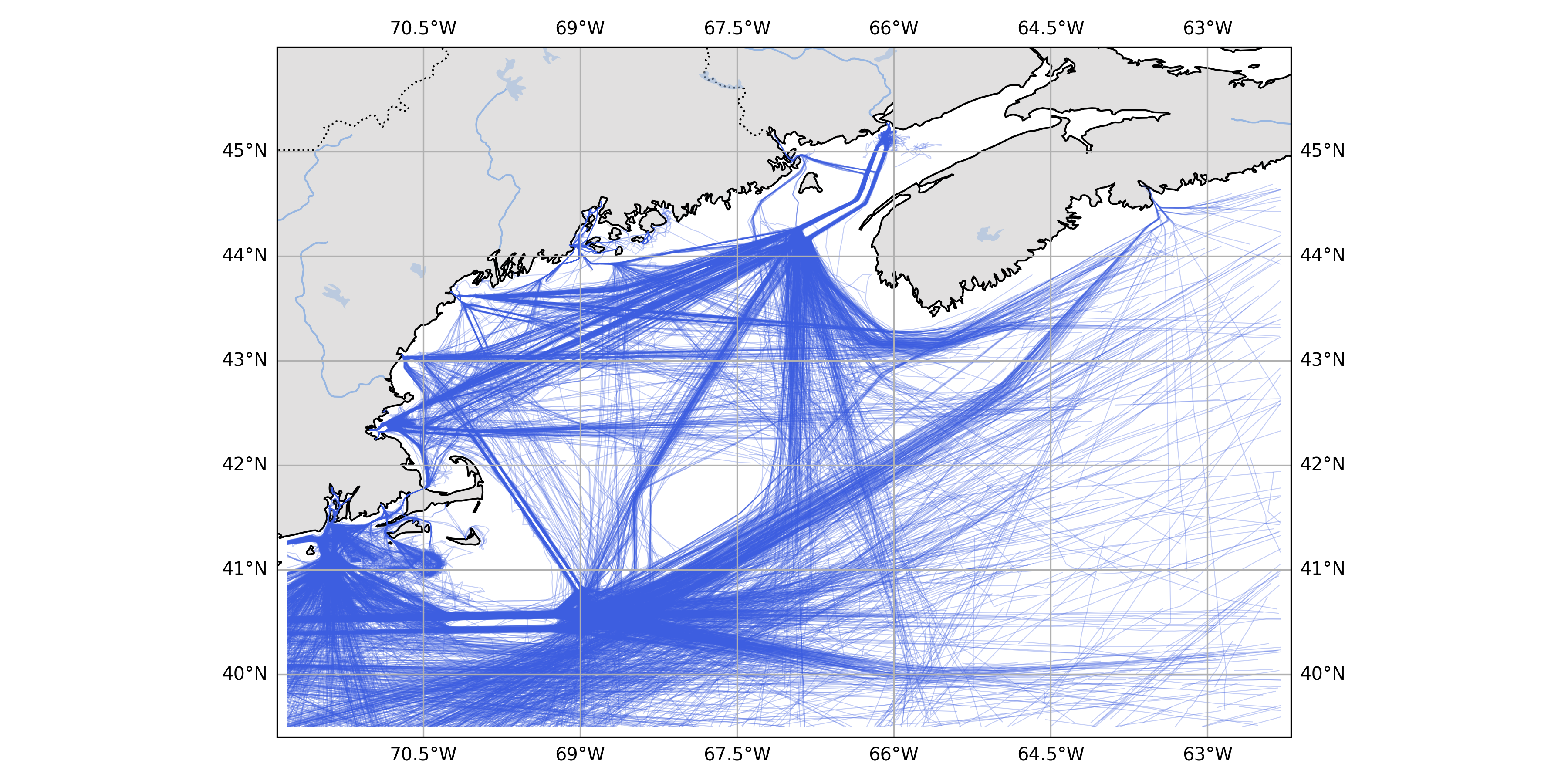}
    \caption{East Coast}
    \label{fig:ec}
\end{subfigure}\hfill
\begin{subfigure}[b]{0.349\linewidth}
    \centering
    \includegraphics[width=\linewidth,trim=95 8 95 8,clip]{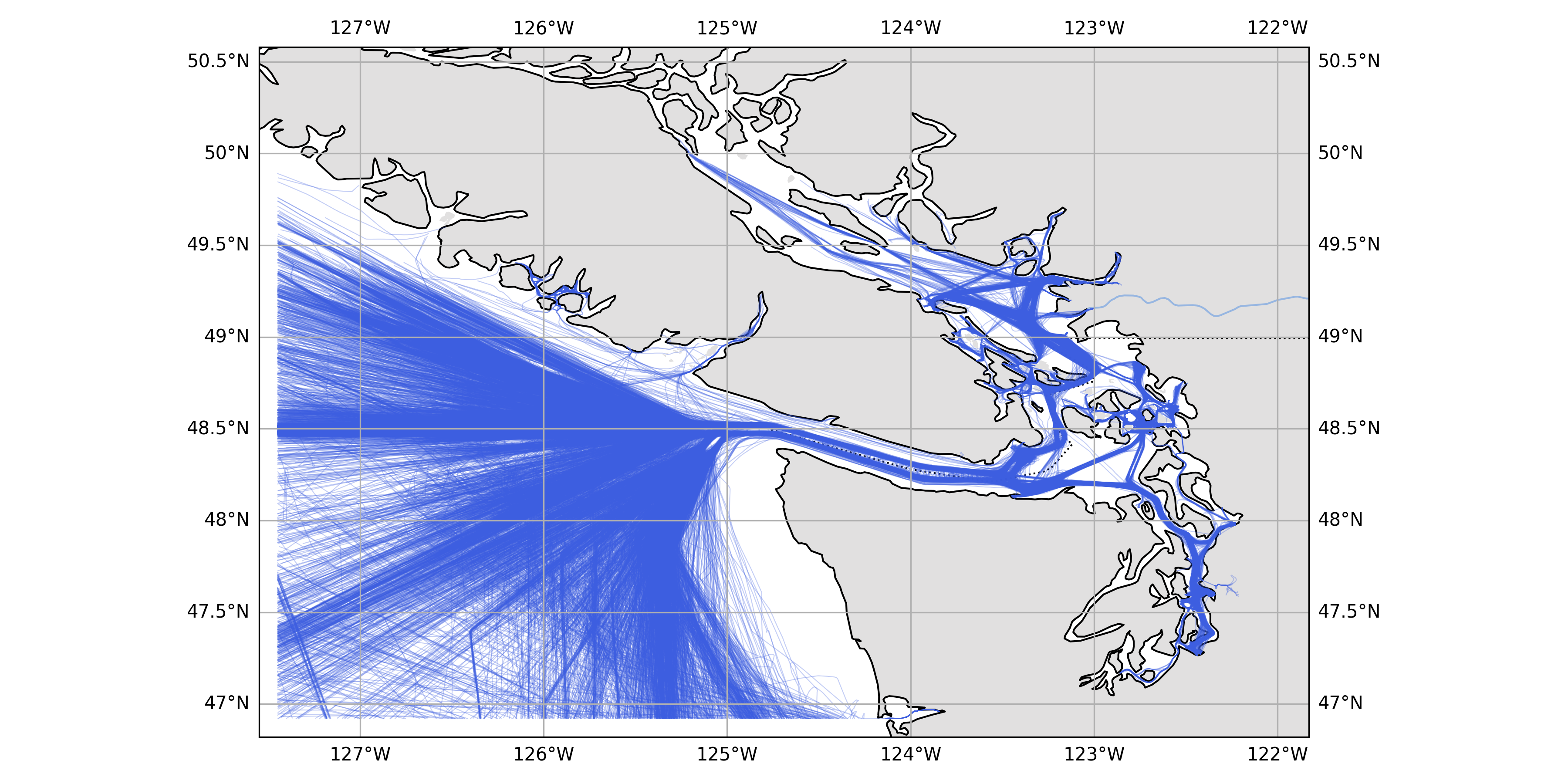}
    \caption{Strait of Georgia}
    \label{fig:sg}
\end{subfigure}\hfill
\begin{subfigure}[b]{0.295\linewidth}
    \centering
    \includegraphics[width=\linewidth,trim=147 8 147 8,clip]{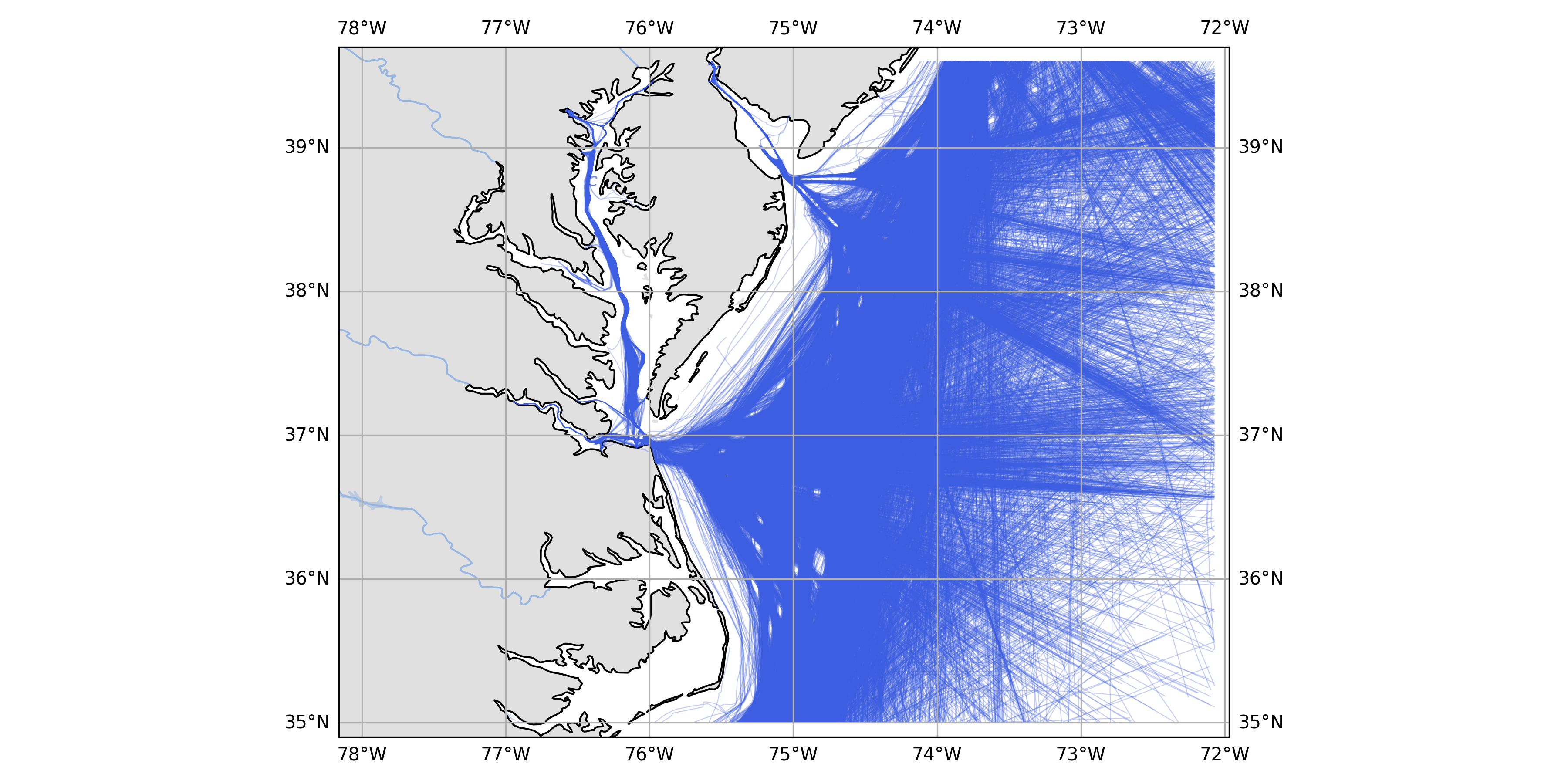}
    \caption{Chesapeake Bay}
    \label{fig:cb}
\end{subfigure}
\caption{Geographical coverage of three regional datasets.}
\label{fig:eval_regions}
\end{figure*}

\subsubsection{Datasets and Preprocessing}
\label{subsec:preprocess}
We conduct experiments using AIS data from Marine Cadastre, collected from three regions: the East Coast near South Nova Scotia, Canada and Maine, USA (January--December 2024, LON: $-71.80$ to $-62.30$, LAT: $39.50$ to $45.90$), the Chesapeake Bay on the US east coast (January--December 2024, LON: $-78.06$ to $-72.07$, LAT: $35.00$ to $39.60$), and the Strait of Georgia on the Pacific coast (January--December 2024, LON: $-127.45$ to $-121.93$, LAT: $46.92$ to $50.48$). For each region, AIS messages include the Maritime Mobile Service Identity (MMSI), timestamp, latitude, longitude, speed, course, and ship type. After filtering by ship type and MMSI, messages were grouped by MMSI to form vessel trajectories. The trajectories were then prepared using the following steps:

\begin{itemize}
    \item \textbf{\textit{Noise removal}}: AIS data may appear on land due to transmission or projection errors. We filter such points using coastal and inland water boundaries. Ships operating at very low speeds ($v_t < 0.5$ knots), typically indicating anchoring or loitering, are excluded to avoid clustering of stationary behaviors. Points with speeds exceeding $40$ knots are also removed as potential errors.

    \item \textbf{\textit{Segmentation}}: Since a trajectory may contain multiple trips, it is segmented into individual trips at temporal gaps of more than $1$ hour or spatial gaps exceeding $100{,}000$ meters between consecutive messages.

    \item \textbf{\textit{Interpolation}}: We resample the segmented trajectories at 2-minute intervals using linear interpolation of latitude and longitude with the nearest points.

    \item \textbf{\textit{Length filtering}}: After interpolation, trajectories with fewer than 50 or more than 3,000 points are removed based on the observed length distribution.
\end{itemize}

\noindent
Dataset statistics per region appear in Table~\ref{tab:datasets} and Fig.~\ref{fig:eval_regions}.

\begin{table}[!htbp]
\centering
\caption{Dataset Statistics}
\label{tab:datasets}
\resizebox{\columnwidth}{!}{%
    \begin{tabular}{l|r|r|r|r}
        \hline
        \textbf{Region} & \textbf{\# Vessels} & \textbf{\# Trajectories} & \textbf{\# Points} & \textbf{Area ($\textnormal{nmi}^2$)} \\ \hline
        East Coast        & 3,066 & 8,658  & 4,310,435  & 160,991 \\ \hline
        Chesapeake Bay    & 7,053 & 15,824 & 14,054,482 & 78,992  \\ \hline
        Strait of Georgia & 5,140 & 12,977 & 15,208,276 & 46,747  \\ \hline
    \end{tabular}%
}
\end{table}

\subsubsection{Baseline Methods}
\label{subsec:baselines}
Our baselines were two distance-based metrics (Hausdorff, DTW) and two learning-based methods (t2vec~\cite{Li2018DeepRepresentationLearning} and TrajCL~\cite{Chang2023ContrastiveTrajectorySimilarity}). We evaluated Bi-LSTM~\cite{Schuster1997BidirectionalRecurrentNeural}, Bi-GRU~\cite{Cho2014LearningPhraseRepresentations}, and TCN~\cite{Lea2016TemporalConvolutionalNetworks} backbones.

\noindent\textbf{\textit{Baseline methods.}} \noindent\textbf{\textit{Baseline methods.}} We evaluate against the published implementations of \emph{t2vec} and \emph{TrajCL}, using their respective pretrained cell representations and original training objectives. For \emph{t2vec}, we learn a spatially aware cell embedding table using skip-gram over the training split and optimize the original objective with the $k$-nearest-cell Kullback--Leibler divergence term. For \emph{TrajCL}, the structural stream uses cell embeddings pretrained by random walks over the cell adjacency graph, with the authors' released hyperparameters: learning rate $10^{-3}$, MoCo momentum $0.999$, and cell and sequence embedding width 256. Under each method's published retrieval protocol on the Strait of Georgia data (train 6K, evaluate 1K, 20\% query sampling), t2vec achieves a mean rank of 3.955, while TrajCL achieves 4.607. Two deviations from the original settings remain due to hardware constraints: the batch size is 16 instead of 128, and the original 200-point length cap is not applied, as maritime cell sequences average 313 tokens and attention cost grows quadratically with sequence length.

\subsubsection{Hyperparameter Settings}
\label{subsubsec:hyperparameter_tuning}
\hfill\par

\textbf{\textit{MoCo-AIS hyperparameters:}} \textit{GRU} and \textit{LSTM} encoders used 2 bidirectional layers to avoid gradient instability in deeper stacks. The \textit{TCN} encoder used a depth of 4, kernel size 3, and a base dilation of 1. A uniform dropout of $0.1$ and the hidden dimension $d_{model}=128$ were used in all encoders.

Models were trained for up to $600$ epochs with batch sizes of $32$--$128$, using the \textit{Adam} optimizer with a learning rate between $0.0001$--$0.001$, adaptively reduced using \textit{ReduceLROnPlateau} when validation performance stagnated. Early stopping with patience $10$--$25$ epochs prevented overfitting.

The momentum update coefficient (\textit{i.e.}, $m$ in (\ref{eq:momentum-update})) was set within $0.99$--$0.9999$ to stabilize key encoder drift~\cite{He2020MomentumContrastUnsupervised, Grill2020BootstrapYourOwn}. The memory queue size, $256$--$1024$, exceeded the batch size to ensure sufficient negatives. The InfoNCE loss used a temperature parameter $\tau$ in the range $0.01$--$0.07$.

\textbf{\textit{Baseline hyperparameters:}} For \textit{t2vec}, we used a two-layer bidirectional GRU with embedding and hidden dimensions $128$, and dropout $0.1$. The model was optimized with \textit{Adam}, a learning rate of $0.001$, and gradient clipping at norm $1.0$. The model uses the original composite objective: the noise-contrastive decoding term over the $k=10$ nearest cells of each target ($\lambda$ weighting as published) together with the triplet term. Cell embeddings of dimension $128$ are pretrained with skip-gram over the training split and, as in the original, remain trainable. The cell grid is $0.05^\circ$, chosen so that trajectories span a number of cells comparable to the road-network setting the method was originally designed for.

For \textit{TrajCL}, we follow the authors' released configuration: a Transformer encoder with 2 attention layers, 4 heads, a feed-forward width of 2048, and cell and sequence embedding dimensions both equal to 256; MoCo queue size 2048, temperature $0.05$, momentum $0.999$; \textit{Adam} at learning rate $10^{-3}$ with weight decay $10^{-4}$ and step decay of $0.5$. Cell embeddings are pretrained by random walks over the cell adjacency graph and, as in the original, frozen. Given the larger spatial extent of maritime regions, we increased the cell size from $0.001^\circ$ to $0.01^\circ$ (approximately $1.1$\,km) and scaled the local spatial mask to the same eleven-cell span. We applied a margin $\epsilon = 2.5 \times 10^{-4}$ for robustness against out-of-boundary trajectories. The batch size is 16 rather than the original 128, which the memory budget cannot admit at this sequence length.

\subsubsection{Implementation and Hardware}
\label{subsubsec:hardware}
All experiments run on one workstation: Intel Core i7-8700K CPU (6 cores, 12 threads, 3.70\,GHz), 32\,GB memory, and two GPUs: an NVIDIA GeForce RTX 2080 Ti (11\,GB) and an NVIDIA TITAN Xp (12\,GB). The software stack is Python 3.12, PyTorch 2.5.1 with CUDA 11.8, and driver 560.35. Every runtime in Table~\ref{tab:compute_time} was measured on the RTX 2080 Ti in one session with no other job on the device, so the encoder rows are comparable. Workloads are not ran concurrently, since sharing a GPU distorts both memory headroom and wall-clock time.

Two considerations qualify this comparison. First, the Hausdorff and DTW baselines use single-threaded CPU implementations (\texttt{hausdorff} and \texttt{fastdtw}), so their reported runtimes represent single-core execution. Second, GPU inference excludes the one-time training cost, which is reported separately. Thus, the comparison reflects the recurring cost of distance computation versus the one-time cost of learning an embedding.

\subsubsection{Evaluation Metrics}
We evaluate representation quality using \textit{Mean Rank} (MR) and \textit{Hitting Ratio} (HR). The metric \textit{Mean Rank}~\cite{Li2018DeepRepresentationLearning} measures how effectively the learned embeddings retrieve each trajectory's augmented counterparts at the top of the retrieval list, reflecting the model's ability to cluster semantically similar trajectories in the latent space. Meanwhile, \textit{Hitting Ratio}~\cite{chen2005robust, su2020survey} quantifies alignment between the learned similarity structure and distance-based metrics (\textit{e.g.}, DTW and Hausdorff), evaluating the preservation of neighborhood relationships. Together, these metrics assess both the discriminative quality of the learned embeddings and their consistency with established similarity definitions.

\textbf{\textit{Mean Rank:}} It evaluates the effectiveness of the learned embeddings in capturing trajectory similarity by measuring the rank of augmented trajectories (\textit{i.e.}, the ground truth in our setting) for each query. Candidates are sorted by cosine distance from the query in the embedding space, and the positions of its augmented trajectories define the target ranks. MR is computed as the average target rank across all queries, where $N$ is the number of queried trajectories; lower values indicate that the ground truth is closer to the top. For each trajectory $i$, the target rank $\text{Rank}(i) = rank(\left | i^{+} \right |, D^{+}_i)$ is the average rank of three augmented trajectories $\left | i^{+} \right |$ over the candidate list $D^{+}i$, giving the overall metric:
\begin{equation}
    \label{eq:mean_rank}
    \text{MeanRank} = \frac{1}{N} \sum_{i=1}^{N} \text{Rank}(i)
\end{equation}

\textbf{\textit{Hitting Rate:}} It measures the overlap between the top-$K$ neighbors of a query trajectory retrieved from the embedding space and those obtained using a distance-based metric. For each queried trajectory $i$, let $\mathcal{N}^{\text{embed}}_K(i)$ and $\mathcal{N}^{\text{dist}}_K(i)$ denote the top-$K$ nearest trajectories in the embedding space and under the distance metric, respectively, defined as:
\begin{align}
    \text{HR}@K(i) &= \frac{\left | \mathcal{N}^{\text{embed}}_K(i) \cap \mathcal{N}^{\text{dist}}_K(i) \right |}{K} \\
    \text{HR}@K &= \frac{1}{N} \sum_{i=1}^{N} \text{HR}@K(i)
\label{eq:hit_rate}
\end{align}

The query trajectory is excluded from both neighbor sets to avoid a trivial self-match. Since each query is its own nearest neighbor under both the embedding and reference distances, including it would introduce a guaranteed overlap of one and artificially inflate HR@1 for all methods. To complement the top-$K$ evaluation, we also report the mean per-query Spearman rank correlation $\rho$ between the embedding-based and reference-based rankings over the remaining $N-1$ trajectories. Whereas HR@K measures agreement within the top-$K$ neighborhood, $\rho$ evaluates the consistency of the complete ranking. Importantly, both metrics quantify agreement with a \emph{geometric} reference rather than performance on an independent task. Consequently, a high score indicates that the embedding preserves geometric neighborhood structure, but not that this structure is attributable to representation learning.

\subsection{Results and Analysis}
\label{sec:exp-evaluation}
The evaluation examines trajectory representation from three complementary perspectives. We first assess how well the learned embeddings preserve neighborhood structures defined by conventional geometric similarity measures. We then ask whether they capture route-level semantics through an origin--destination (OD) discrimination task. Finally, we examine cross-regional generalization and efficiency.

\subsubsection{Neighborhood Preservation and Its Controls}
\label{sec:hitting-ratio}
We first evaluate the agreement between the learned embedding space and the neighborhood structure induced by Hausdorff distance. 
We quantify this alignment with HR (\ref{eq:hit_rate}) against the cached Hausdorff reference over \emph{all} evaluation queries, and with the mean per-query Spearman correlation $\rho$, which extends HR@$K$ to the full neighbor ranking. 

Table~\ref{tab:hitrate_controls} reports both together with the reference levels required to read them: the \emph{chance} floor (a random ordering, HR@$K = K/(N-1)$), a \emph{centroid} baseline that discards trajectory shape entirely and compares only mean position, and, for every encoder, the \emph{same architecture at random initialization} averaged over three seeds, which passes input geometry through an unlearned random projection. 

\begin{table}[!htbp]
\caption{Hitting Ratio and rank correlation against the Hausdorff reference for all queries evaluated. For each encoder, the random-init control uses the same architecture without training (mean over three seeds; 
std $\leq 0.016$). All models are trained on 1K trajectories. 
The lower block reports cell-token inputs.}
\label{tab:hitrate_controls}
\centering
\resizebox{\columnwidth}{!}{%
\begin{tabular}{l ccc ccc}
\toprule
\multirow{2}{*}{Method} & \multicolumn{3}{c}{Evaluation size 1K} & \multicolumn{3}{c}{Evaluation size 9K} \\
\cmidrule(lr){2-4}\cmidrule(lr){5-7}
 & HR@1 & HR@10 & $\rho$ & HR@1 & HR@10 & $\rho$ \\
\midrule
Chance & 0.001 & 0.010 & 0.000 & 0.000 & 0.001 & 0.000 \\
Centroid (no shape) & 0.220 & 0.461 & 0.897 & 0.117 & 0.273 & 0.917 \\
\midrule
\multicolumn{7}{l}{\textit{Raw coordinate input}} \\
\quad BiGRU random-init & 0.255 & 0.472 & 0.874 & 0.142 & 0.291 & 0.916 \\
\quad BiGRU MoCo-AIS & \textbf{0.266} & 0.430 & 0.371 & \textbf{0.146} & 0.276 & 0.640 \\
\quad BiLSTM random-init & 0.272 & 0.481 & 0.885 & 0.151 & 0.302 & 0.913 \\
\quad BiLSTM MoCo-AIS & \textbf{0.327} & \textbf{0.527} & 0.217 & \textbf{0.163} & 0.297 & 0.553 \\
\quad TCN random-init & 0.270 & 0.481 & 0.862 & 0.149 & 0.304 & 0.910 \\
\quad TCN MoCo-AIS & \textbf{0.312} & \textbf{0.528} & 0.308 & \textbf{0.153} & 0.295 & 0.491 \\
\midrule
\multicolumn{7}{l}{\revised{\textit{Cell-token input}}} \\
\quad \revised{BiGRU random-init} & \revised{0.310} & \revised{0.444} & \revised{0.188} & \revised{0.251} & \revised{0.373} & \revised{0.165} \\
\quad \revised{BiGRU MoCo-AIS} & \revised{0.272} & \revised{0.396} & \revised{0.261} & \revised{0.243} & \revised{0.335} & \revised{0.246} \\
\quad \revised{t2vec (restored)} & \revised{0.242} & \revised{0.395} & \revised{0.542} & \revised{0.222} & \revised{0.316} & \revised{0.557} \\
\bottomrule
\end{tabular}%
}
\end{table}

Every method agrees with the Hausdorff reference far more than chance, showing that the learned embedding spaces preserve meaningful geometric neighborhood structure. For raw-coordinate inputs, all three MoCo-AIS encoders improve HR@1 over their random-init controls on the 1K evaluation set, with BiLSTM achieving the highest HR@1 of 0.327. This agreement remains substantial across evaluation sizes, indicating that the learned representations capture stable neighborhood structure rather than isolated top-ranked matches. The cell-token results further demonstrate the benefit of spatial
discretization. By mapping consecutive observations to shared spatial cells, tokenization reduces fine-grained positional variation while retaining coarse trajectory structure. 

\subsubsection{Downstream Evaluation: Origin--Destination Route Discrimination}
\label{sec:od-downstream}
To evaluate the embeddings against a ground truth external to the training objective, we derive \emph{OD route labels} from the data without manual annotation. The first and last points of all 9K evaluation trajectories are pooled and clustered with DBSCAN (haversine, $\varepsilon = 5$\,km, \texttt{min\_samples} $= 5$) into port zones, and zones whose bounding box exceeds 50\,km are discarded as chained corridors rather than ports. Each label is the \emph{ordered} pair of origin and destination zones, yielding 1{,}657 labeled trajectories in 642 directed classes. Two tasks are evaluated under this labeling. In \emph{retrieval}, every labeled trajectory whose class has at least two members (1{,}256 queries) queries the full labeled pool, and a retrieved trajectory counts as relevant if and only if it shares the query's OD class; we report HR@$K$, mean reciprocal rank (MRR), and mean average precision (mAP). In \emph{clustering}, average-linkage agglomerative clustering is applied to each method's distance matrix over the classes with at least five members (825 trajectories, $k=63$) and scored by normalized mutual information (NMI), adjusted Rand index (ARI), and purity. The endpoint baseline uses the same endpoint information used to define the OD labels and is therefore included only as an upper bound.

\begin{table}[!htbp]
\caption{Downstream OD route discrimination on the 9K evaluation set with directed labels. All learned models are trained on 1K trajectories. For \emph{retrieval}, HR@$K$ is the fraction of 1{,}256 queries with a same-OD trajectory among the top $K$ neighbors. For \emph{clustering}, hierarchical clustering with average linkage was applied to each method's distance matrix over 825 trajectories from 63 OD classes. \emph{Input}: raw coordinates or grid-cell tokens. Bold denotes the best result; the endpoint baseline is an upper bound: it uses the label-defining information. Random-init is averaged over three seeds.}
\label{tab:od_downstream}
\centering
\resizebox{\columnwidth}{!}{%
\begin{tabular}{llccccccc}
\toprule
\multirow{2}{*}{Method} & \multirow{2}{*}{Input} & \multicolumn{4}{c}{Retrieval} & \multicolumn{3}{c}{Clustering} \\
\cmidrule(lr){3-6}\cmidrule(lr){7-9}
 & & HR@1 & HR@10 & MRR & mAP & NMI & ARI & Purity \\
\midrule
Chance (permutation) & --   & 0.010 & 0.091 & 0.041 & 0.014 & -- & -- & -- \\
Centroid             & --   & 0.464 & 0.796 & 0.580 & 0.432 & 0.789 & 0.530 & 0.611 \\
Hausdorff            & --   & 0.689 & \textbf{0.937} & 0.784 & \textbf{0.641} & \textbf{0.855} & 0.581 & \textbf{0.682} \\
\midrule
Random-init BiGRU    & raw  & 0.545 & 0.803 & 0.632 & 0.446 & 0.780 & 0.476 & 0.581 \\
\revised{Random-init BiGRU} & \revised{cell} & \revised{0.716} & \revised{0.881} & \revised{0.774} & \revised{0.595} & \revised{0.815} & \revised{0.521} & \revised{0.606} \\
\revised{t2vec (restored)}  & \revised{cell} & \revised{\textbf{0.760}} & \revised{0.902} & \revised{\textbf{0.811}} & \revised{0.639} & \revised{0.820} & \revised{0.509} & \revised{0.613} \\
MoCo-AIS (BiGRU)     & raw  & 0.603 & 0.833 & 0.680 & 0.469 & 0.748 & 0.390 & 0.543 \\
\revised{MoCo-AIS (BiGRU)}  & \revised{cell} & \revised{0.742} & \revised{0.897} & \revised{0.796} & \revised{0.637} & \revised{0.843} & \revised{\textbf{0.591}} & \revised{0.672} \\
\rowcolor{gray!15}
Endpoint (OD oracle) & --   & 0.974 & 1.000 & 0.985 & 0.968 & 0.982 & 0.959 & 0.953 \\
\bottomrule
\end{tabular}%
}
\end{table}

On retrieval, MoCo-AIS consistently outperforms its architecture-matched random initialization on both input representations, improving HR@1 from 0.545 to 0.603 with raw coordinates and from 0.716 to 0.742 with cell tokens. This improvement holds on OD labels never used during training, evidence that the learned representation transfers to an external route-discrimination task. With cell tokens, MoCo-AIS also approaches the best retrieval performance among learned methods, achieved by t2vec (0.742 vs.\ 0.760).

The benefit of spatial discretization is clearer in clustering. MoCo-AIS improves from an ARI of 0.390 with raw coordinates to 0.591 with cell tokens, the highest non-oracle ARI in the table. Cell tokens reduce fine-grained positional variation while preserving the spatial structure shared by trajectories on the same route, giving substantially better separation of OD classes. The clustering results therefore give complementary evidence that the representation captures route-level structure beyond geometric similarity alone.

The directed OD evaluation further shows that sequence-based representations retain travel direction. When directed labels are replaced by unordered labels, geometric methods improve, whereas all sequence-based models lose accuracy, with MoCo-AIS decreasing by 0.023 and 0.034 for raw and cell-token inputs, respectively. This behavior indicates that sequential representations capture directional information that is not represented by symmetric geometric distances.

\begin{figure}[!htbp]
    \centering
    \includegraphics[width=\columnwidth]{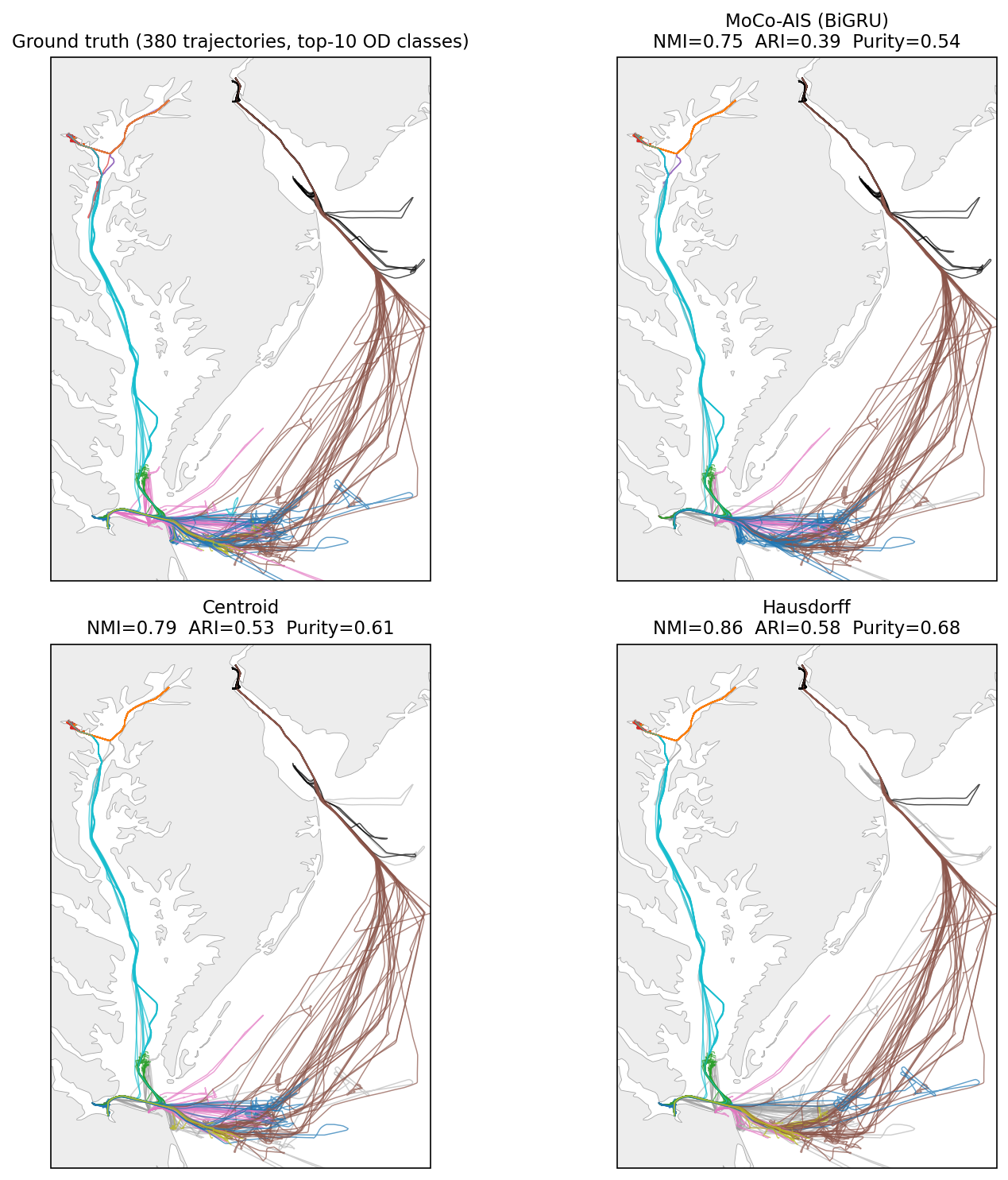}
    \caption{OD route clustering on the 9K evaluation set, showing the top-10 directed OD classes. Each cluster is assigned the ground-truth OD class of its majority members and displayed using that class's color; trajectories assigned to a different OD class are shown in gray.}
    \label{fig:od_clustering_map}
\end{figure}

\subsubsection{Representation-Aligned Comparison with Learning-based Baselines}
\label{sec:baseline-aligned}
Both learning-based baselines use grid-cell token sequences, whereas the default MoCo-AIS configuration uses raw coordinate sequences. Directly comparing these models would therefore conflate the effects of the input representation with those of the learning framework. To isolate these effects, we evaluate MoCo-AIS using the same cell grid, vocabulary, and pretrained cell embeddings as t2vec, while keeping the remaining MoCo-AIS pipeline unchanged. For all tokenized models, the vocabulary and cell embeddings are constructed only from the training split. All models in Table~\ref{tab:baseline_aligned} are trained on the same 9K trajectories and evaluated on a disjoint 1K set.

\begin{table}[!htbp]
\caption{OD route retrieval with fixed input representation (directed labels). All learned models are trained on the same 9K split and evaluated on a disjoint 1K test set. Bold marks the best non-oracle entry per column; the endpoint baseline is a leakage bound (gray). Random-init is averaged over 3 seeds.}
\label{tab:baseline_aligned}
\centering
\resizebox{\columnwidth}{!}{%
\begin{tabular}{llcccc}
\toprule
Method & Input & HR@1 & HR@10 & MRR & mAP \\
\midrule
Chance (permutation) & --   & 0.036 & 0.281 & 0.114 & 0.064 \\
Centroid             & --   & 0.719 & 0.942 & 0.800 & 0.723 \\
Hausdorff            & --   & 0.884 & 0.983 & 0.921 & 0.848 \\
\midrule
Random-init BiGRU    & raw  & 0.722 & 0.942 & 0.804 & 0.724 \\
Random-init BiGRU    & cell & 0.848 & 0.978 & 0.903 & 0.830 \\
\midrule
t2vec (restored)     & cell & 0.868 & \textbf{0.992} & 0.919 & \textbf{0.850} \\
TrajCL (restored)    & cell & 0.793 & 0.934 & 0.841 & 0.665 \\
\midrule
MoCo-AIS (BiGRU)     & raw  & 0.835 & 0.975 & 0.884 & 0.799 \\
MoCo-AIS (BiGRU)     & cell & \textbf{0.934} & \textbf{0.992} & \textbf{0.953} & 0.810 \\
\rowcolor{gray!15}
Endpoint (OD oracle) & --   & 1.000 & 1.000 & 1.000 & 0.999 \\
\bottomrule
\end{tabular}%
}
\\[2pt]
{\footnotesize This split contains 121 eligible queries; the binomial 95\% confidence half-width near HR@1 = 0.9 is approximately $\pm0.053$. Clustering is not reported because only 69 trajectories belong to OD classes with at least five members, which is insufficient for a stable comparison. Clustering results are therefore reported only for the larger evaluation set in Table~\ref{tab:od_downstream}.}
\end{table}

Input representation has a strong effect on retrieval performance.
With the framework held fixed, replacing raw coordinates with cell tokens
increases HR@1 from 0.835 to 0.934, while the corresponding random-init
control improves from 0.722 to 0.848. This improvement indicates that
spatial discretization itself provides a representation better suited to the
OD retrieval task. By mapping nearby observations to shared cells,
discretization suppresses fine-grained positional variation between different
transits of the same route while preserving their shared spatial structure.

With the input representation aligned, our framework achieves the
highest HR@1 among the learned models in Table~\ref{tab:baseline_aligned},
reaching 0.934 compared with 0.868 for t2vec and 0.793 for TrajCL. It also
outperforms the Hausdorff baseline (0.884) on this evaluation split. These
results show that the benefit of cell-token input is retained when the
learning framework is compared under the same representation.

The random-init cell-token control achieves an HR@1 of 0.848 without
training, while contrastive learning further improves it to 0.934. Thus,
cell-tokenization provides a strong representation for OD retrieval, and
contrastive learning provides an additional improvement on top of it.

\subsubsection{Retrieval and Robustness under Perturbation Protocols}
\label{sec:mean-rank}
Retrieval is also evaluated under the augmentation-based protocol of prior work~\cite{Li2018DeepRepresentationLearning}, in which the perturbed copies of a trajectory serve as the ground truth. Because those perturbations belong to the families used to build the training pairs, the protocol measures invariance to the training augmentations rather than similarity in any external sense. We therefore report it only in the form that uses perturbation families that have not been applied during training, which is the version that carries information about the representation.

Table~\ref{tab:heldout} reports this evaluation by perturbing each E1 trajectory and ranking its original trajectory among all $N=1{,}000$ candidates. For the shifting family, which is included during training, MoCo-AIS achieves a mean rank (MR) of 1.000. On the three held-out perturbation families, however, the corresponding random-initialized encoder performs better, with MR values of 2.23 versus 1.19 for downsampling, 1.21 versus 1.15 for masking, and 60.7 versus 24.8 for simplification. Hausdorff achieves MR values of 1.04--1.45 without training. Because each perturbed trajectory retains most of the geometry of its original, high retrieval performance can be obtained directly from geometric similarity. Thus, MR primarily measures invariance to the perturbations rather than the quality of the learned representation.

\begin{table}[!htbp]
\centering
\caption{Mean rank (MR, $\downarrow$) when a \emph{perturbed} copy of each trajectory queries the pool of $N = 1{,}000$ originals (E1 split). Shifting and sub-trajectory are used to construct training pairs, while downsampling, masking, and simplification are held out during training. A random ranking has an expected mean rank of 500.5; random-init reports the mean rank over three seeds.}
\label{tab:heldout}
\resizebox{\columnwidth}{!}{%
\begin{tabular}{l rrrr}
\toprule
Method & Shifting$^\dagger$ & Downsample & Mask & Simplify \\
\midrule
MoCo-AIS (BiGRU)  & 1.000 & 2.229 & 1.211 & 60.73 \\
BiGRU random-init & 1.003 & 1.188 & 1.146 & 24.75 \\
Hausdorff         & 1.000 & \textbf{1.045} & \textbf{1.037} & \textbf{1.454} \\
\bottomrule
\end{tabular}%
}
\\[2pt]
{\footnotesize $^\dagger$Training augmentation family; other columns are held out from training.}
\end{table}

To evaluate cross-regional generalization, we trained models on the East Coast (EC) 6K dataset and assessed retrieval on the Chesapeake Bay (CB) 6K dataset and the Strait of Georgia (SG) 3K and 9K datasets (Table~\ref{tab:cross_region_rank}). Cross-regional retrieval is consistently more challenging than same-region evaluation, and the drop is most evident for the TCN encoder, whose rank percentage rises to 2.7\% on the Strait of Georgia against 0.15\% on Chesapeake Bay, suggesting reduced adaptability under regional shifts. All encoders transfer better to Chesapeake Bay, the region geographically closer to the training data, indicating that the learned representations capture region-specific mobility characteristics that benefit nearby regions but limit broader generalization. Rank percentages nevertheless remain small, and the recurrent encoders achieve the lowest values in both evaluation areas, indicating greater robustness to regional distribution shifts than the TCN encoder.

\begin{table}[!htbp]
\centering
\small
\setlength{\tabcolsep}{3pt}
\caption{Mean rank (MR, $\downarrow$) and rank percentage (\%, $\downarrow$) under cross-region evaluation. Training on 6K trajectories of the East Coast (EC); evaluation sizes are 6K from Chesapeake Bay (CB) and 3K and 9K from the Strait of Georgia (SG).}
\label{tab:cross_region_rank}
\resizebox{\columnwidth}{!}{%
\begin{tabular}{c l rrrrrr}
\toprule
\multirow{2}{*}{Train} & \multirow{2}{*}{Method}
  & \multicolumn{2}{c}{6K (CB)}
  & \multicolumn{2}{c}{3K (SG)}
  & \multicolumn{2}{c}{9K (SG)} \\
\cmidrule(lr){3-4} \cmidrule(lr){5-6} \cmidrule(lr){7-8}
  & & MR & (\%) & MR & (\%) & MR & (\%) \\
\midrule
\multirow{3}{*}{6K (EC)}
  & BiGRU       & 16.102 & 0.268 & \textbf{3.832} & \textbf{0.128} & \textbf{9.440} & \textbf{0.105} \\
  & BiLSTM      & \textbf{5.375} & \textbf{0.090} & \underline{7.619} & \underline{0.254} & \underline{21.902} & \underline{0.243} \\
  & TCN         & 9.208 & 0.153 & 80.924 & 2.697 & 239.009 & 2.656 \\
\bottomrule
\end{tabular}%
}
\end{table}

\subsubsection{Computational Efficiency}
\label{sec:efficiency}

To evaluate the computational efficiency, we logged the time required for three distinct cases: (1) the similarity computation by traditional distance-based methods, (2) the inference procedure for embedding the evaluation data into the learned space, and (3) the computation of cosine distance matrices based on the obtained embeddings.

Given the substantially higher computational cost of traditional distance-based metrics, the runtimes reported for Hausdorff and DTW in Table~\ref{tab:compute_time} are expressed in hours. A clear contrast in scalability emerges between these methods and learning-based models. Hausdorff and DTW exhibit quadratic growth in runtime, since the similarity matrix requires $N(N-1)/2$ pairwise computations, increasing from several hours at smaller evaluation sizes to nearly 77 hours for Hausdorff at larger scales. For DTW, the 9K runtimes were prohibitively long and are therefore omitted; program estimates indicate that the 9K setting would exceed one month.

\begin{table}[!htbp]
\centering
\caption[Computational time comparison]{Computational time of distance-based \textit{vs.} learning-based methods. For distance-based metrics, the total time to compute the similarity matrix is reported in \textit{hours}. For learning-based methods, \textit{Train} is the one-off training cost on the 1K training set and times include embedding inference (\textit{Infr.}) and cosine distance matrix computation (\textit{Dist.}), in \textit{seconds}. All values were measured on a single RTX 2080 Ti with no concurrent job (Section~\ref{subsubsec:hardware}); the distance-based baselines are single-threaded CPU implementations.}
\label{tab:compute_time}

\resizebox{\columnwidth}{!}{%
\begin{tabular}{l r rr rr rr rr}
\toprule
\multirow{2}{*}{Methods} & \multirow{2}{*}{\revised{Train}} & \multicolumn{2}{c}{1K} & \multicolumn{2}{c}{3K} & \multicolumn{2}{c}{6K} & \multicolumn{2}{c}{9K} \\
\cmidrule(lr){3-4}\cmidrule(lr){5-6}\cmidrule(lr){7-8}\cmidrule(lr){9-10}
\multicolumn{10}{c}{\hspace{1.8cm}\textit{Distance-based metrics (total time only, in hours)}} \\
Hausdorff & \revised{--} & \multicolumn{2}{c}{1.04} & \multicolumn{2}{c}{8.77} & \multicolumn{2}{c}{34.39} & \multicolumn{2}{c}{77.66} \\
DTW       & \revised{--} & \multicolumn{2}{c}{5.49} & \multicolumn{2}{c}{49.44} & \multicolumn{2}{c}{190.85} & \multicolumn{2}{c}{--} \\
\midrule
\multicolumn{10}{c}{\hspace{1.8cm}\textit{MoCo-AIS with different encoders (time in seconds)}} \\
& & Infr. & Dist. & Infr. & Dist. & Infr. & Dist. & Infr. & Dist. \\
\cmidrule(lr){3-4}\cmidrule(lr){5-6}\cmidrule(lr){7-8}\cmidrule(lr){9-10}
\revised{BiGRU}       & \revised{711} & \revised{1.53} & \revised{0.001} & \revised{3.67} & \revised{0.014} & \revised{7.00} & \revised{0.047} & \revised{10.11} & \revised{0.082} \\
\revised{BiLSTM}      & \revised{564} & \revised{1.46} & \revised{0.002} & \revised{3.66} & \revised{0.014} & \revised{6.77} & \revised{0.037} & \revised{10.10} & \revised{0.099} \\
\revised{TCN}         & \revised{579} & \revised{2.40} & \revised{0.001} & \revised{6.38} & \revised{0.014} & \revised{13.24} & \revised{0.053} & \revised{19.73} & \revised{0.099} \\
\bottomrule
\end{tabular}%
}
\\[2pt]
{\footnotesize Training early stopping epochs: 165 (BiGRU), 128 (BiLSTM), 100 (TCN).}
\end{table}

\begin{table}[!htbp]
\centering
\caption{Training and inference time for the learning-based methods, all
trained on the 9K split and measured on an RTX 2080 Ti with no concurrent load. Training is a one-time cost; inference is the time to embed the evaluation set.}
\label{tab:baseline_compute}
\centering
\small
\setlength{\tabcolsep}{5pt}
\begin{tabular}{llrrr}
\toprule
\multirow{2}{*}{Method} & \multirow{2}{*}{Input} & \multicolumn{1}{c}{Train} & \multicolumn{2}{c}{Inference (s)} \\
\cmidrule(lr){3-3}\cmidrule(lr){4-5}
 & & \multicolumn{1}{c}{(s)} & \multicolumn{1}{c}{1K} & \multicolumn{1}{c}{9K} \\
\midrule
t2vec (restored)  & cell & 5374 & \textbf{0.19} & \textbf{1.10} \\
TrajCL (restored) & cell & 5084 & 4.27 & 29.01 \\
MoCo-AIS (BiGRU)  & raw  & 2400 & 1.53 & 10.11 \\
MoCo-AIS (BiGRU)  & cell & \textbf{536}  & 0.76 & 2.98 \\
\bottomrule
\end{tabular}
\\[2pt]
{\footnotesize Inference time measures the cost of converting the evaluation trajectories into embeddings, including preprocessing and the model forward pass.}
\end{table}

Table~\ref{tab:baseline_compute} compares the computational cost with the same 9K training size. Because training schedules differ across methods, we compare their cost by wall-clock time rather than epoch count. Cell-token MoCo-AIS requires 536\,s to train, compared with 5374\,s for t2vec and 5084\,s for TrajCL, and reduces inference time at 9K from 10.11\,s to 2.98\,s via shorter sequences. t2vec remains the fastest at inference (1.10\,s), and the advantage of MoCo-AIS over it is in training efficiency.

In contrast, embedding inference for learning-based encoders scales linearly with the size of the evaluation set, whereas pairwise distance computation scales quadratically. At 9K trajectories, GPU-based cosine-distance computation takes less than 0.1\,s, compared with hours for single-threaded geometric baselines, yielding a speedup of more than $10^4\times$. Among the learned encoders, TCN is the slowest at inference (19.7\,s at 9K), while the recurrent encoders require about 10.1\,s. The trained encoder can be reused for new trajectory datasets, whereas pairwise distance methods must recompute all trajectory-to-trajectory distances for each dataset.

\section{Conclusion}
\label{sec:conclusion}

We presented MoCo-AIS, a unified contrastive framework for learning vessel trajectory embeddings with interchangeable sequence encoders. Our results show that learned embeddings preserve meaningful trajectory structure and transfer to an external origin--destination route discrimination task. More importantly, the input representation has a stronger effect than the choice of encoder: spatial cell tokens substantially improve route-level discrimination, while contrastive learning further refines the resulting representation. Learned embeddings also provide a scalable alternative to pairwise trajectory distances, reducing large-scale similarity computation from hours to seconds. These findings underscore representation design as a central factor in effective and scalable trajectory similarity learning. Although designed for maritime data, both the framework and the control-based evaluation protocol apply to other mobility domains, such as urban traffic and human mobility analysis, where similar spatiotemporal challenges arise.

The remaining challenge is the MoCo-style pair construction for large-scale free-space trajectories: augmentation-based positives emphasize intra-trajectory invariance, while sampled negatives may include trajectories on the same routes. Future work can add lightweight supervision to filter semantically related negatives and build more informative positives.

\section*{Acknowledgment}
\sloppy
This research was partially supported by the \textit{Natural Sciences and Engineering Research Council} (NSERC RGPIN-2025-05179), the \textit{National Council for Scientific and Technological Development} (CNPq 444325/2024-7), and the Faculty of Computer Science at \textit{Dalhousie University} (DAL). The AIS data used in this study were obtained from Marine Cadastre, an open-source ocean data portal. The code and data are available on Figshare: \url{https://figshare.com/s/189382cd16eef9cf074f}

\section*{Disclosure, Ethics, and Responsible AI Use}
This study uses pre-collected vessel-traffic records, involves no human subjects, and does not intervene in marine operations. The authors declare no competing interests. Generative AI technologies were used to assist with drafting and code scaffolding; the authors are responsible for verifying all claims, analyses, figures, and results before submission.

\bibliographystyle{IEEEtran}
\bibliography{references_mocoais}

\end{document}